\documentclass[letterpaper, 10 pt, journal, twoside]{IEEEtran}

\usepackage{multirow}
\usepackage{threeparttable}
\usepackage{bm}
\usepackage{svg}
\usepackage{amsmath}
\usepackage[normalem]{ulem}
\usepackage{cite}
\usepackage{makecell}
\usepackage[subrefformat=parens]{subcaption}

\title{
Soft and Rigid Object Grasping With Cross-Structure Hand Using Bilateral Control-Based Imitation Learning
}

\author{
    Koki Yamane$^{1}$, Yuki Saigusa$^{1}$, Sho Sakaino$^{1}$ and Toshiaki Tsuji$^{2}$
    \thanks{Manuscript received: July 8, 2023; Revised October 3, 2023; Accepted October 25, 2023.}
    \thanks{This paper was recommended for publication by Editor Clement Gosselin upon evaluation of the Associate Editor and Reviewers’ comments. This work was supported by NEDO and JSPS.} 
    \thanks{
        $^{1}$The authors are with the Graduate School of Systems and Information Engineering, University of Tsukuba, 1-1-1 Tennodai, Tsukuba, Ibaraki 305-8577, Japan
        {\tt\small yamane.koki.td@tsukuba.alumni.ac.jp}
    }
    \thanks{
        $^{2}$The author is with the Graduate School of Science and Engineering, Saitama University, Saitama 338-8570, Japan
    }%
    \thanks{Digital Object Identifier (DOI): see top of this page.}
}

\begin{document}

\markboth{IEEE Robotics and Automation Letters. Preprint Version. 11, 2023}
{Yamane \MakeLowercase{\textit{et al.}}: Grasping With Cross-Structure Hand Using Bilateral Control-Based Imitation Learning} 

\maketitle

\begin{abstract}

Object grasping is an important ability required for various robot tasks.
In particular, tasks that require precise force adjustments during operation, such as grasping an unknown object or using a grasped tool, are difficult for humans to program in advance.
Recently, AI-based algorithms that can imitate human force skills have been actively explored as a solution.
In particular, bilateral control-based imitation learning achieves human-level motion speeds with environmental adaptability, only requiring human demonstration and without programming.
However, owing to hardware limitations, its grasping performance remains limited, and tasks that involves grasping various objects are yet to be achieved.
Here, we developed a cross-structure hand to grasp various objects.
We experimentally demonstrated that the integration of bilateral control-based imitation learning and the cross-structure hand is effective for grasping various objects and harnessing tools.

\end{abstract}

\section{INTRODUCTION}

\IEEEPARstart{R}{obots} are expected to replace human physical labor, and object grasping is an important skill required for various tasks.
In particular, tasks that require precise force adjustments during operation, such as grasping an unknown object or using a grasped tool, are difficult for humans to program in advance.

Accordingly, methods that employ machine learning have been proposed to perform such complex tasks.
Reinforcement learning, which is one of such methods, allows agents to be trained by simply defining a reward function.
Mart{\'\i}n-Mart{\'\i}n \textit{et al.} performed reinforcement learning using variable impedance control in an end-effector space (VICES) as the action space, and successfully completed the task with significant contact with the robot~\cite{martin2019variable}.
Beltran-Hernandez \textit{et al.} achieved a contact-rich task using modified parallel position as well as force and admittance control as the action space for reinforcement learning~\cite{beltran2020learning}.
However, a few challenges persist, such as the difficulty in defining an appropriate reward function for the task and the need for many trials owing to low sample efficiency.
Furthermore, several methods employ simulation to generate large amounts of data at a low cost; however, it is difficult to simulate non-rigid objects.

In contrast, imitation learning, which learns behaviors from human demonstrations, is known as a method with high sample efficiency because it learns from successful data.
With imitation learning, a human can acquire complex behaviors simply by providing instructional data requiring neither difficult programming nor the definition of a reward function.
Force-controlled methods have been proposed, especially for contact-intensive tasks.
Rozo \textit{et al.} adapted Gaussian mixture models and variable impedance control to achieve human-robot coordination tasks~\cite{rozo2015learning}.
Wang \textit{et al.} combined trajectory imitation learning and reinforcement learning-based force control to achieve assembly tasks~\cite{wang2021robotic}.
However, the teaching data collection method often used in imitation learning, namely direct teaching~\cite{kormushev2011imitation}, in which the robot is directly manipulated by human touch, makes it difficult to record the force applied by the human because forces applied by the human and reaction forces from the environment are applied to the robot simultaneously and cancel each other out.
In addition, direct teaching records the observed values of each joint angle, and imitation learning adopts the recorded values as the correct data for the command values; however, in high-speed movements, there is an error between the command and observed values owing to the delay of the control system.
Therefore, most imitation learning methods require the robot to move sufficiently slow for the delay to be negligible.
Consequently, the robot moves significantly slower than humans.

However, bilateral control-based imitation learning~\cite{adachi2018imitation} was proposed as an adaptive force control system.
In 4-channel bilateral control~\cite{sakaino2011multi}, two robots, a leader and a follower, are employed to measure the reaction force applied by the robot from the environment and that applied by the human separately.
By adopting the teaching data of the force, adaptive motion similar to that of humans can be achieved via continuous motion determination and fast motion.
This method has been validated in human-robot cooperation~\cite{sasagawa2020imitation}, letter writing~\cite{hayashi2022independently}, and scoops and transportation of pancakes~\cite{saigusa2022imitation}.
However, these experiments were conducted with a tool fixed to the robot, and bilateral control-based imitation learning is yet to be validated for tasks including grasping, such as grasp of various unknown objects or use of a grasped tool, owing to the constraints imposed by the ability and operability of the hardware, including grippers. 

Although various grippers have been proposed for object grasp, few are suitable for measuring human force.
Conventional grippers are based on suction or one-degree-of-freedom rotary and linear motion types~\cite{zhang2020state};  however, simple structured grippers can only grasp a limited range of objects.
Various types of grippers that can grasp several hard and soft objects have been proposed.
For example,
two-finger tendon-driven grippers~\cite{hussain2018modeling}, which are composed of rigid links connected through flexible joints;
jamming grippers~\cite{amend2012positive}, which adaptively change their shapes by controlling the air pressure in a balloon membrane filled with grains;
and farmHand~\cite{ruotolo2021grasping}, which generates van der Waals forces using the structure of a gecko's finger.
Nevertheless, because these grippers are made of soft materials, it is difficult for humans to manipulate them by applying forces directly from the outside, and it is also difficult to measure these forces because neither forces applied by humans nor reaction forces from the environment are transmitted to the actuator.
To record movements including applied and reaction forces, it is required that a gripper features a rigid body that can be easily manipulated with a low degree of freedom and can grasp various objects.

In this study, we developed a gripper suitable for object grasping using bilateral control and successfully collected teaching data for object grasping tasks that require force levels to be fine-tuned during operation, such as grasping several unknown objects and employing a grasped tool.
The developed gripper can grasp various objects, including small and thin objects, owing to its crossed structure, while satisfying the requirements in 4-channel bilateral control.
It solely includes rigid bodies and can be easily manipulated with one degree of freedom.
In addition, the developed gripper was subjected to imitation learning with teaching data obtained using the developed gripper and achieved high success rates in two tasks: grasping various unknown objects which required soft grasping, and a letter-writing task, which required rigid grasping.

The key contributions of this study are presented below.
\begin{itemize}
    \item We developed a simple single-degree-of-freedom hand with cross-structure that can measure the force of a human motion in complex grasping tasks.
    \item We experimentally verified that the robot can perform soft and rigid grasping in two tasks: grasping various unknown objects and using a grasped tool.
\end{itemize}

\section{Related Works}

\subsection{Object Grasping}

Object grasping is an important ability required in several robot tasks; hence, most of this study was conducted accordingly.
Existing models can be broadly divided into two types: model-based methods, in which the 3D shape and category of the object are known, and model-free methods, in which there is no prior information about the object.
The model-based method provides a highly reliable grasp.
However, it cannot guarantee action on unknown objects~\cite{kleeberger2020survey}. 
In contrast, model-free methods can handle unknown objects.
Consequently, model-free methods are expected to be implemented.
A model-free method has been proposed to train a neural network using RGB-D images or point clouds as input, and a graspable pose has been adopted as the output
~\cite{fang2020graspnet,zeng2021transporter}.
Although these methods can handle objects of various shapes, they require a large amount of data for generalization because they take high-dimensional data as input.
In addition, it is difficult to consider intrinsic properties such as the density and stiffness of an object based solely on visual information.

\subsection{Imitation Learning}

Imitation learning, which learns motions from human manipulation data, has been garnering attention as a method with high sample efficiency among robot control methods using machine learning, and successful learning of grasping motions has been reported~\cite{argall2009survey,rahmatizadeh2016learning,mandlekar2021matters}.
It is an end-to-end learning method that learns the required output directly from the data and can easily handle complex behaviors to reproduce following a human trajectory.


\subsection{Bilateral Control-Based Imitation Learning}

Bilateral control-based imitation learning
~\cite{adachi2018imitation,fujimoto2019time,sasagawa2020imitation,hayashi2022independently,saigusa2022imitation} 
is an imitation learning method that employs data collected by 4-channel bilateral control.

In direct teaching, which is the typical approach to collect teacher data for imitation learning, a human applies a force to a robot directly, and the robot is subjected to both the human-applied force and the reaction force from the environment simultaneously, which cannot be measured independently.
Meanwhile, 
4-channel bilateral control employs two robots and separates the robot on the side where the human applies the force and the robot on the side where the robot receives the reaction force from the environment.
It allows obtaining information on the human-applied force and the reaction force independently.
Using this information for imitation learning allows mimicking the force applied by humans.

Furthermore, bilateral control-based imitation learning can move robots at the same speed as humans.
Although in conventional imitation learning, the next response value is predicted from the current response value, the response and command values do not match because robots exhibit phase delay due to imperfect control.
Therefore, robots need to move at a speed that allows them to ignore this difference and move significantly slower than humans.
Meanwhile, bilateral control-based imitation learning predicts the next command value based on the response value of the autonomous robot.
It enables learning that considers the control delay and achieves fast autonomous motion like humans.

\section{Control System}

\subsection{Manipulator}
\begin{figure}[!t]
    \centering
    \includegraphics[height=5cm]{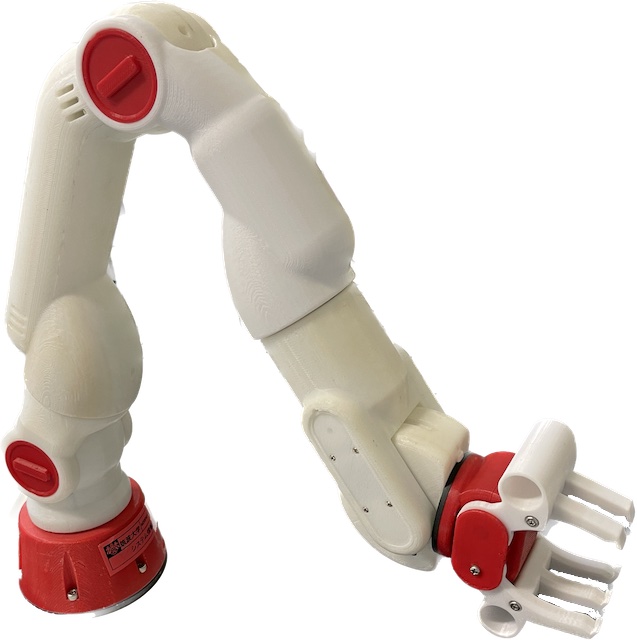}
    \caption{CRANE-X7 with developed hand}
    \label{fig:CRANE-X7}
\end{figure}

CRANE-X7, a manipulator manufactured by RT corporation, was employed.
The appearance of the manipulator is presented in Fig.~\ref{fig:CRANE-X7}.
The manipulator exhibits seven degrees of freedom, while the gripper exhibits one degree of freedom, thereby providing a total of eight degrees of freedom.
We replaced the robot's hand with a hand later described.

\subsection{Position and Force Hybrid Controller}
\begin{figure}[!t]
    \centering
    \includegraphics[width=\linewidth]{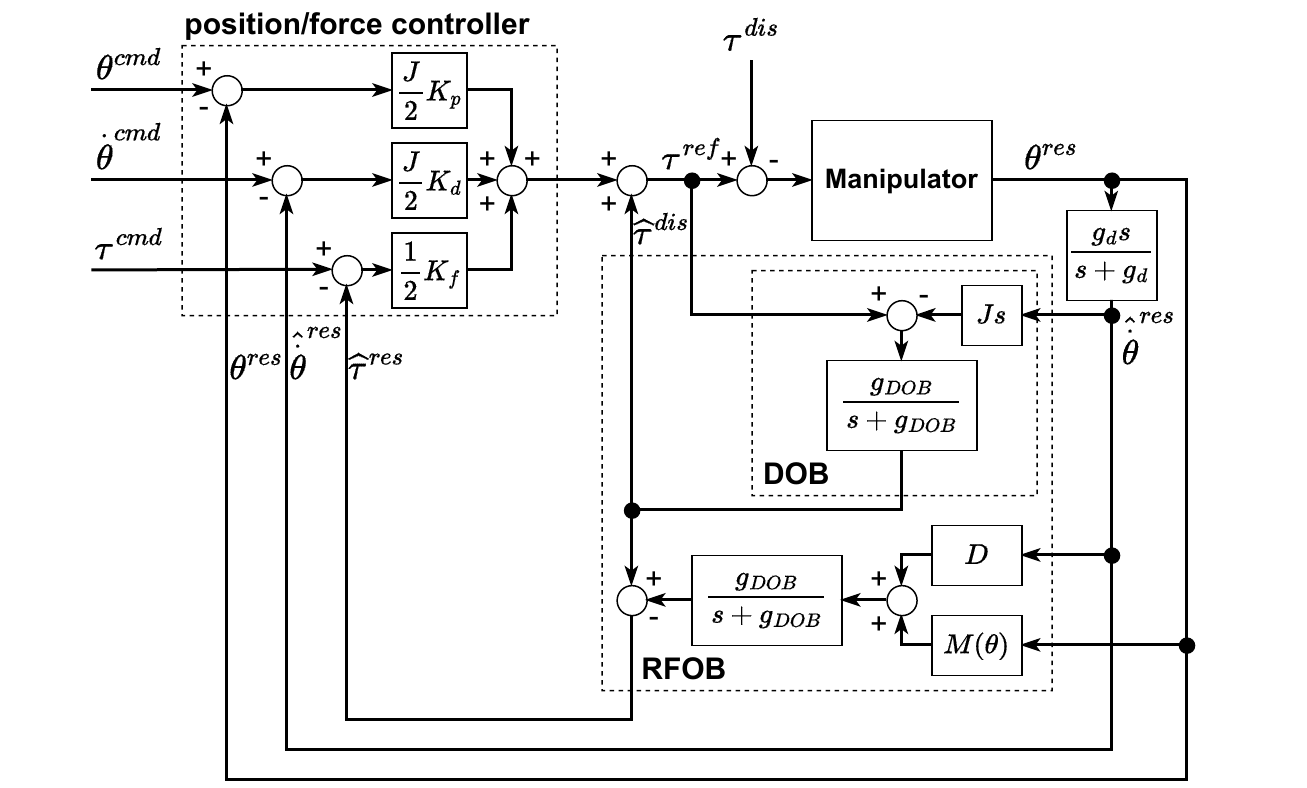}
    \caption{Block diagram of position and force hybrid controller}
    \label{fig:controller}
\end{figure}

Each axis was controlled using a position and force hybrid controller. 
The block diagram of the controller is presented in Fig.~\ref{fig:controller}.

The proportional-derivative (PD) control was applied to the position, and the proportional (P) control was applied to the force. 
The disturbance torque of each joint $\hat{\tau}_{dis}$ was calculated and suppressed using a disturbance observer (DOB)~\cite{ohnishi1996motion}.
The control frequency was set to 500 Hz. 

The angle of each joint was obtained from the rotary encoder, and the angular velocity was calculated by pseudo-differential.
Although the reaction torque applied to each axis needs to be observed for this control, it cannot be measured directly because CRANE-X7 does not include a torque sensor.
Hence, the control system was implemented using sensorless force control.
The torque response value $\hat{\tau}_{res}$ was estimated using a reaction force observer (RFOB)~\cite{murakami1993torque}. 

The torque reference value was calculated as
\begin{align}
    \bm{\tau}^{ref} &= \quad\frac{\bm{J}}{2}\bm{K}_p(\bm{\theta}^{cmd}-\bm{\theta}^{res}) \linebreak \notag \\
        &\quad+\frac{\bm{J}}{2}\bm{K}_d(\bm{\dot{\theta}}^{cmd}-\bm{\hat{\dot{\theta}}}^{res}) \linebreak \notag \\
        &\quad+\frac{1}{2}\bm{K}_f(\bm{\tau}^{cmd}-\bm{\hat{\tau}}^{res}) + \bm{\hat{\tau}}^{dis} \\
    \bm{\hat{\tau}}^{dis} &= \frac{\bm{g}_{DOB}}{s+\bm{g}_{DOB}}(\bm{\tau}^{ref} - \bm{J}s\bm{\hat{\dot{\theta}}}^{res}) \\
    \bm{\hat{\tau}}^{res} &= \bm{\hat{\tau}}^{dis} - \frac{\bm{g}_{DOB}}{s+\bm{g}_{DOB}}(\bm{D}\bm{\hat{\dot{\theta}}}^{res} + \bm{M}(\bm{\theta})) \\
    \bm{\hat{\dot{\theta}}}^{res} &= \frac{\bm{g}_ds}{s+\bm{g}_d}\bm{\theta}^{res}
\end{align}
where $\bm{J}$, $\bm{D}$, $\bm{M}(\theta)$, $\bm{g}_{DOB}$,  and $\bm{g}_d$ are the moment of inertia of the joint, coefficient of viscous friction, gravity, and cut-off frequencies of the low-pass filters, respectively.
In addition, $\bm K_p$, $\bm K_d$, and $\bm K_f$ represent the control gains of position, velocity, and force controls, respectively.
These parameters are set to the same values as presented in related work~\cite{saigusa2022imitation}.
Here, $\theta$, $\dot{\theta}$, and $\tau$ denote the angle, angular velocity, and torque of each joint, respectively, and the superscripts ``cmd,'' ``res,'' ``ref,'' and ``dis'' denote the command, response, reference, and disturbance values, respectively;
$\hat{}$ denotes the estimated value.

\subsection{4-Channel Bilateral Control}

To collect demonstration data for imitation learning, we employed 4-channel bilateral control~\cite{sakaino2011multi}, which is a system that comprises two robots: a leader directly manipulated by a human and a follower of the leader.
The system controls the position and torque of the two robots to synchronize them.
The angle, angular velocity, and torque command values for the 4-channel bilateral control are defined as
\begin{align}
    \bm{\theta}^{cmd}_f &= \bm{\theta}^{res}_l,
    \bm{\theta}^{cmd}_l = \bm{\theta}^{res}_f \\
    \bm{\dot{\theta}}^{cmd}_f &= \bm{\dot{\theta}}^{res}_l, 
    \bm{\dot{\theta}}^{cmd}_l = \bm{\dot{\theta}}^{res}_f \\
    \bm{\tau}^{cmd}_f &= -\bm{\tau}^{res}_l, 
    \bm{\tau}^{cmd}_l = -\bm{\tau}^{res}_f
\end{align}
where subscripts $l$ and $f$ represent the leader and follower, respectively.

Using torque synchronization, the operating human can observe the reaction force from the environment to the follower robot and control the force of the follower robot. 
The position is controlled such that the leader's and follower's joint angles match each other by setting the follower's and leader's joint angles as command values, respectively. 
Regarding the torque, the sign of the torque applied to the leader's joints ($\bm{\tau}^{res}_l$) was reversed to represent the torque exerted by humans, and the sign of the torque applied to the follower's joints ($\bm{\tau}^{res}_f$) was reversed to represent the reaction force received by the follower from the environment.

\section{Cross-structure Hand}

\begin{figure}[!t]
    \centering
    \begin{minipage}[b]{0.49\linewidth}
        \centering
        \includegraphics[width=\linewidth]{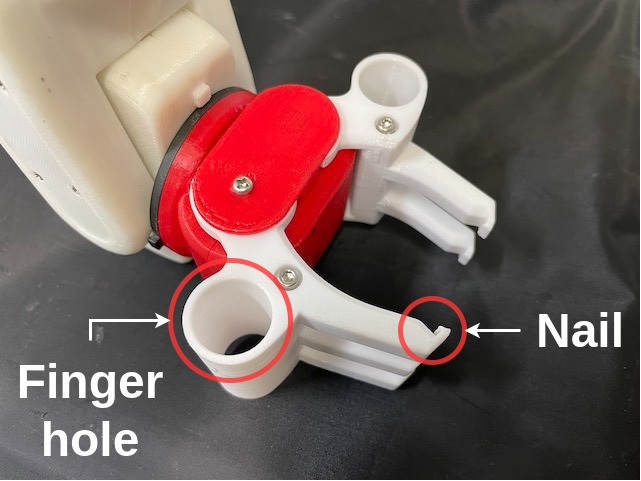}
        \subcaption{opened} 
    \end{minipage}
    \begin{minipage}[b]{0.49\linewidth}
        \centering
        \includegraphics[width=\linewidth]{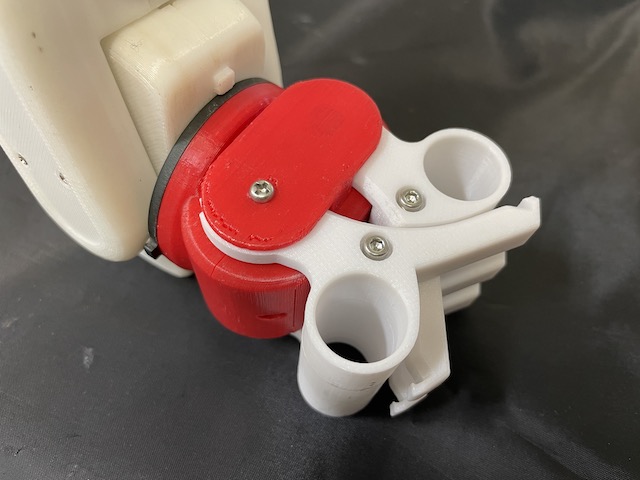}
        \subcaption{closed} 
    \end{minipage}
    \caption{Overview of cross-structure hand}
    \label{fig:hand}
\end{figure}

A simple robot hand was developed for grasping tasks using bilateral control-based imitation learning.
An overview of the hand is shown in Fig.~\ref{fig:hand}.

\begin{figure}[!t]
    \centering
    \begin{minipage}[b]{0.49\linewidth}
        \centering
        \includegraphics[width=\linewidth]{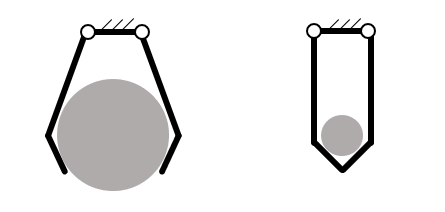}
        \subcaption{typical rotary hand} 
    \end{minipage}
    \begin{minipage}[b]{0.49\linewidth}
        \centering
        \includegraphics[width=\linewidth]{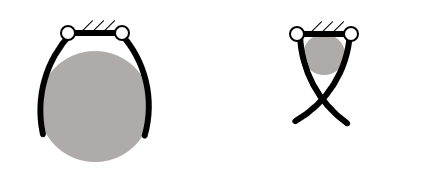}
        \subcaption{cross-structure hand} 
    \end{minipage}
    \caption{Feature of cross-structure hand}
    \label{fig:hand-description}
\end{figure}

Given that robot hands with multiple joints increase the size and cost of the entire mechanism, a two-finger gripper, which has the simplest configuration, is practically preferable~\cite{zhang2020state}.
For a two-finger gripper to enclose an object, the tips of the fingers must be curved inward.
However, with such a shape, the hand cannot be closed any further when the tips of the fingers of the hand hit the tips of the fingers on the opposite side, as illustrated in Fig.~\ref{fig:hand-description}~(a).
This creates a gap on the root side of the hand, and force cannot be applied to an object smaller than this gap.
Therefore, we adopted a shape in which the tip of the hand is divided.
This allows the tip to cross and close to the root, allowing it to grasp thin objects and apply force, as illustrated in  Fig.~\ref{fig:hand-description}~(b).

\begin{figure}[!t]
    \centering
    \begin{minipage}[b]{0.49\linewidth}
        \centering
        \includegraphics[width=\linewidth]{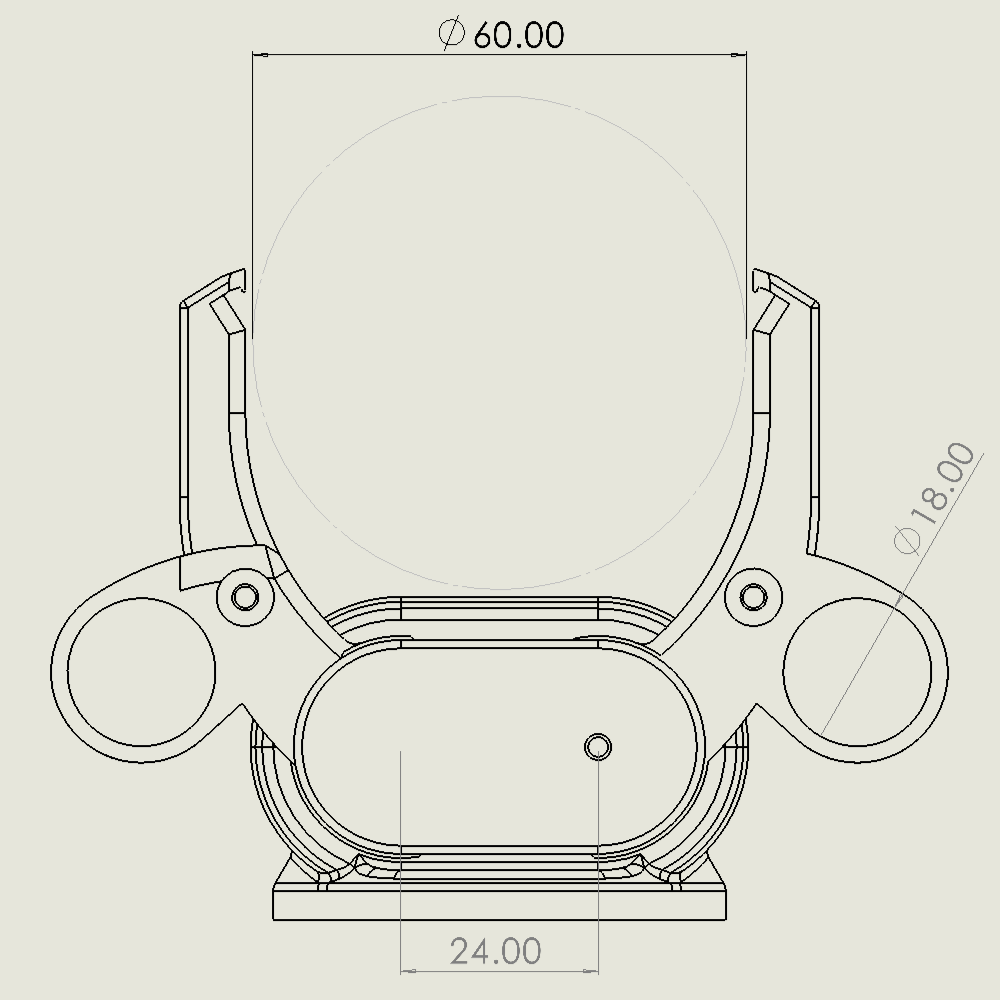}
        \subcaption{maximum grasp size} 
    \end{minipage}
    \begin{minipage}[b]{0.49\linewidth}
        \centering
        \includegraphics[width=\linewidth]{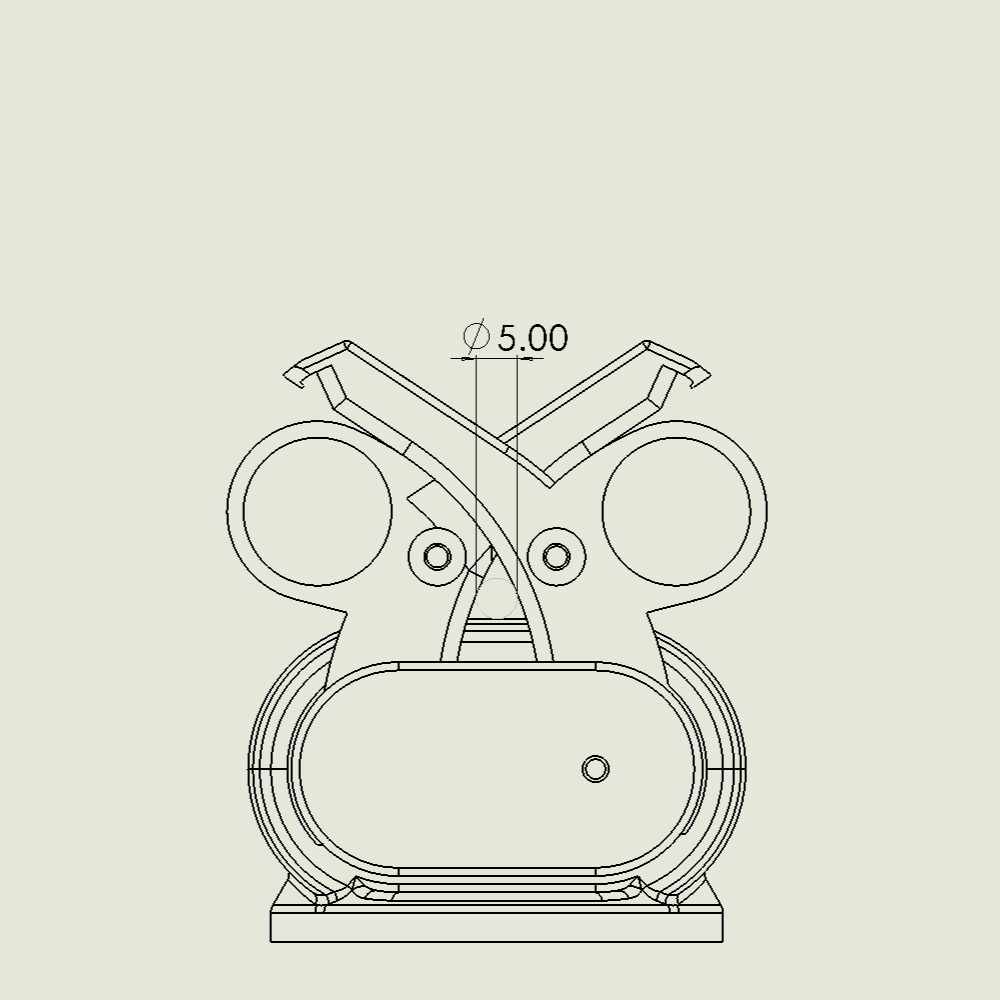}
        \subcaption{minimum grasp size} 
    \end{minipage} \\
    
    \begin{minipage}[b]{0.49\linewidth}
        \centering
        \includegraphics[width=\linewidth]{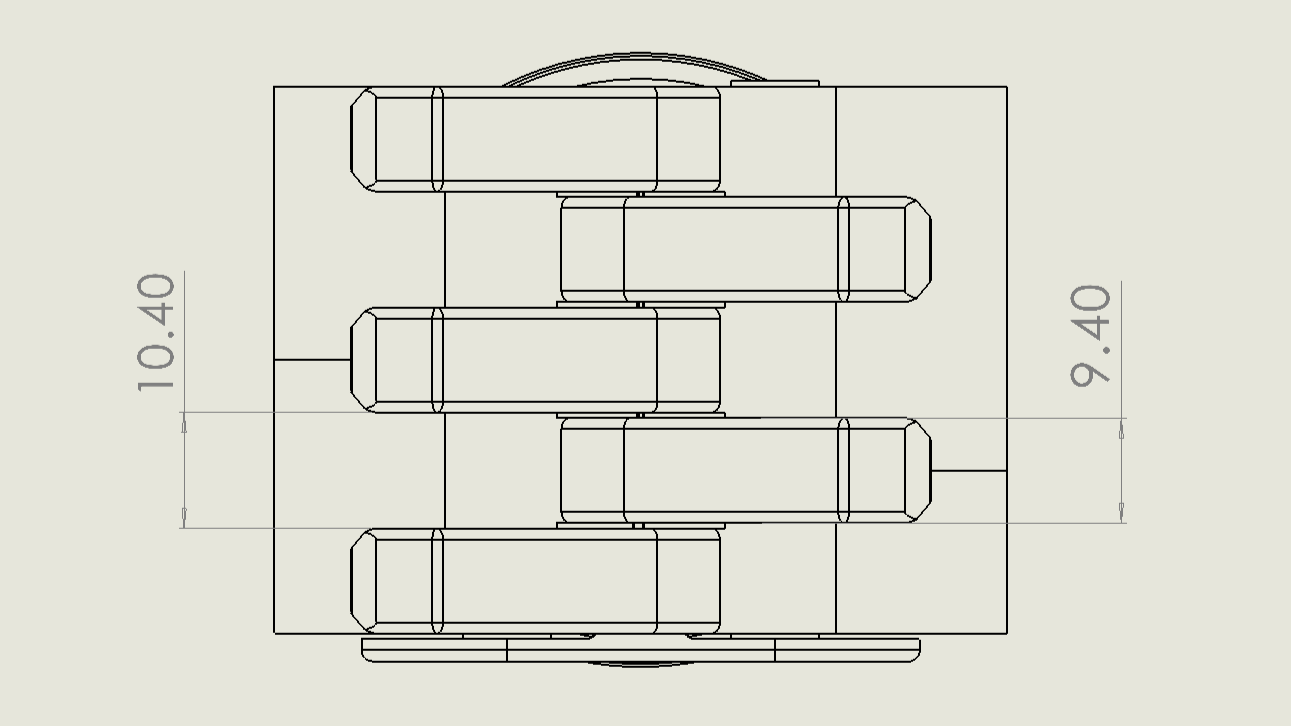}
        \subcaption{finger width and clearance} 
    \end{minipage}
    \begin{minipage}[b]{0.49\linewidth}
        \centering
        \includegraphics[width=\linewidth]{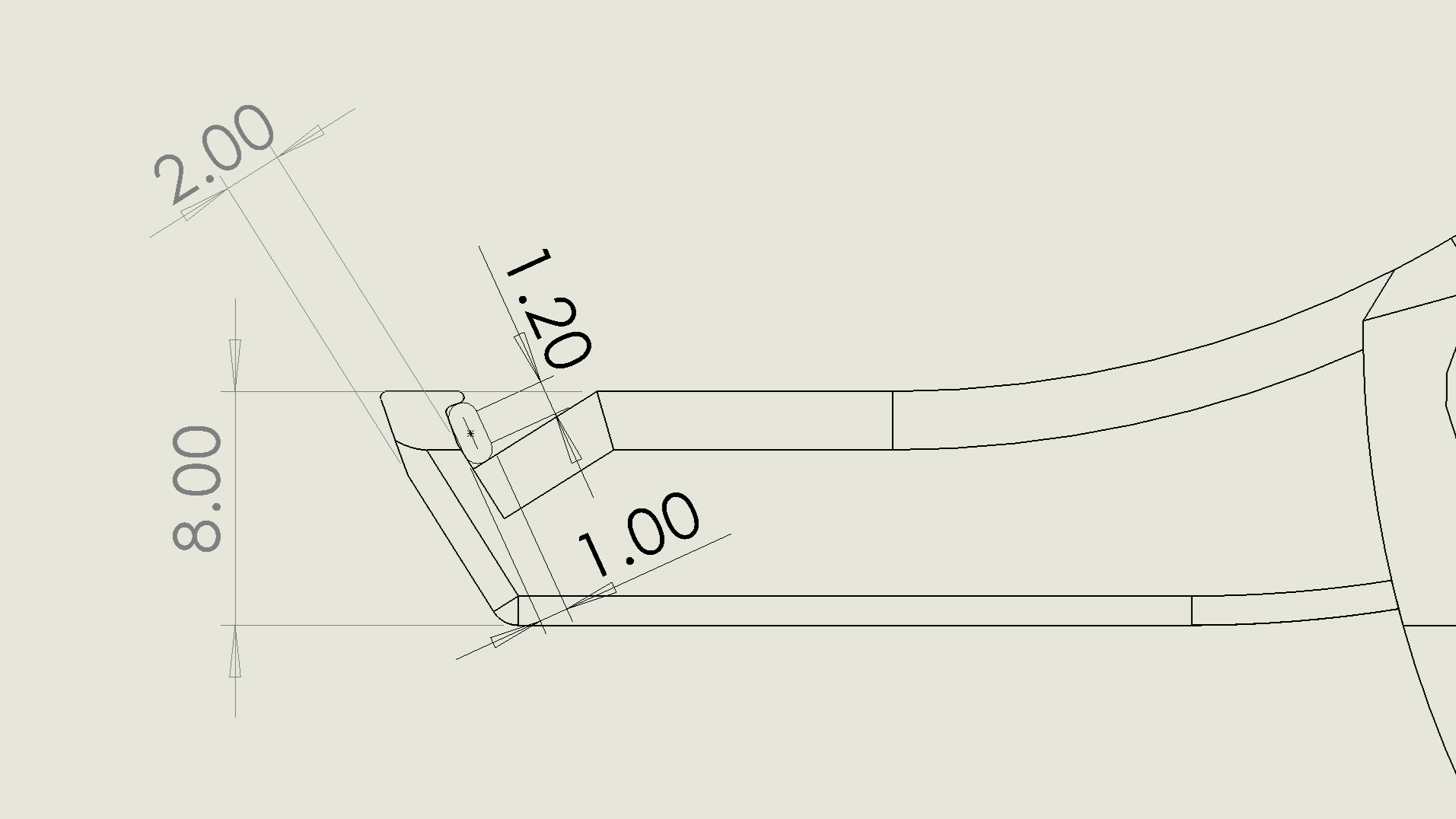}
        \subcaption{nail design} 
    \end{minipage}
    
    \caption{Design of cross-structure hand}
    \label{fig:hand_cad}
\end{figure}

As a design specification, a cylinder with a diameter ranging from 5 to 60 mm was assumed to be grasped.
The maximum and minimum grasp size of the hand are presented in Figs.~\ref{fig:hand_cad}~(a) and ~\ref{fig:hand_cad}~(b).
Moreover, considering the trade-off between strength and graspable size, the finger width was assumed to be 9.4 mm, and the finger space was 10.4 mm (the clearance was 1 mm).
The finger width and space are presented in Fig.~\ref{fig:hand_cad}~(c).
An FDM 3D printer was employed to create this hand using polycarbonate, which is a hard and light element.

In addition, the tip of the hand has a pointed shape similar to that of a nail.
This makes it easier to insert the fingers into the gap between the object and ground when grasping the object and allows strong grasping action at the root side.
The shape of the nail was designed so that the indentation is tangent to the straight slot, as shown in Fig.~\ref{fig:hand_cad}~(d).

Moreover, finger holes were added to improve operability because the robot was intended for bilateral control.
Hence, it is better to operate the entire robot with one hand.


\section{System for Bilateral Control-based Imitation Learning}

\subsection{Network Architecture}

\begin{figure}[!t]
    \centering
    \includegraphics[width=\linewidth]{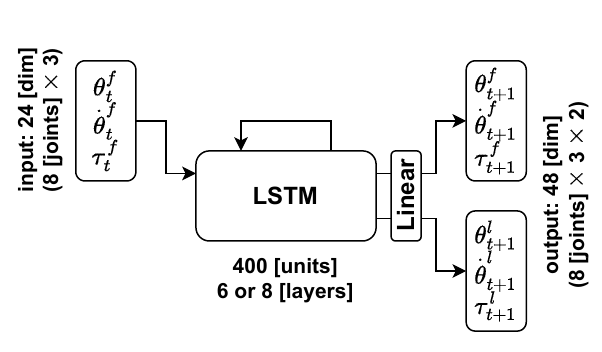}
    \caption{Network architecture}
    \label{fig:LSTM_model}
\end{figure}

Here, the Follower to Follower and Leader (F2FL) model~\cite{hayashi2022independently}, which predicts the next follower and leader states from the current follower state and comprises a long short-term memory (LSTM)~\cite{hochreiter1997long}, was adopted as the neural network architecture for imitation learning.
The proposed network features 24 dimensions of input ([angle, angular velocity, torque] $\times$  [8 degrees of freedom]), 6 or 8 layers of LSTM with 400 units, and one layer including all connected layers with 48 units ([angle, angular velocity, torque] $\times$  [8 degrees of freedom] $\times$  [leader, follower]).
As an output layer, a linear layer was used to connect from 400 dimension LSTM output to the 48-dimension output.
The number of LSTM layers was modified according to the task.
To prevent overtraining, each layer of the LSTM employed dropout, which ignores node connection with a probability of 0.1.
The inputs were the response values of the follower for the current step, whereas the outputs were the leader's and follower's command values for the next step.
The entire picture of the neural network model is illustrated in Fig.~\ref{fig:LSTM_model}.

\subsection{Dataset}

Here, data acquired at 500 Hz were decimated to 25 Hz via sampling every 20 steps.
When training a neural network with time-series data, the learning efficiency can be improved by reducing the sampling frequency to a certain degree~\cite{rahmatizadeh2016learning}.
In addition, the number of data can be increased by a factor of 20.
Furthermore, in combination with a low-pass filter, high-frequency components that do not require imitation can be eliminated from the data, thereby increasing the stability of the operation.
The data were normalized for training the neural network by setting the mean to zero and the standard deviation to one.
In addition, normally distributed noise with a variance of 0.01 was added to the input.

\begin{figure}[!t]
    \centering
    \begin{minipage}[t]{0.32\linewidth}
        \centering
        \includegraphics[width=\linewidth]{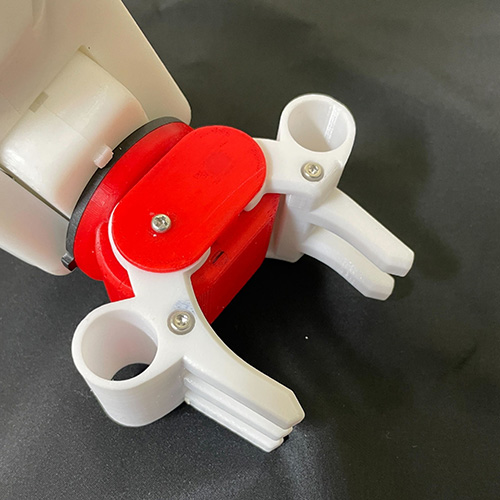}
        \subcaption{w/o nail} 
    \end{minipage}
    \begin{minipage}[t]{0.32\linewidth}
        \centering
        \includegraphics[width=\linewidth]{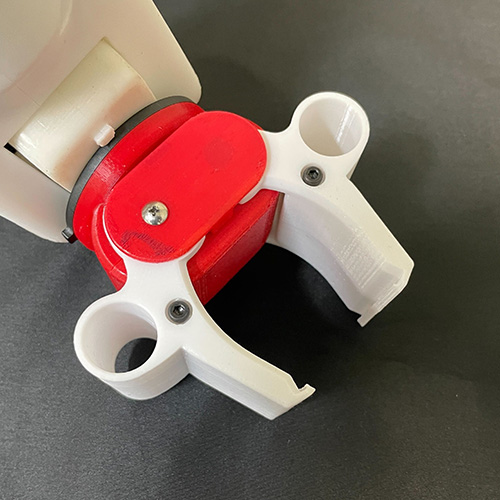}
        \subcaption{w/o cross} 
    \end{minipage}
    \begin{minipage}[t]{0.32\linewidth}
        \centering
        \includegraphics[width=\linewidth]{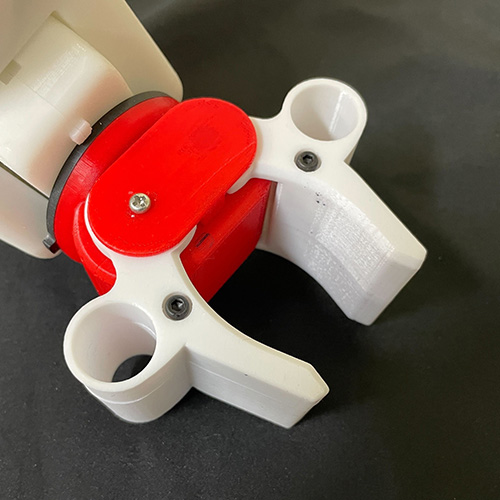}
        \subcaption{w/o nail, cross} 
    \end{minipage}
    \caption{Hands for preliminary experiment}
    \label{fig:hands_for_preliminary_experiments}
\end{figure}

\begin{figure}[!t]
    \centering
    \begin{minipage}[t]{0.49\linewidth}
        \centering
        \includegraphics[width=\linewidth]{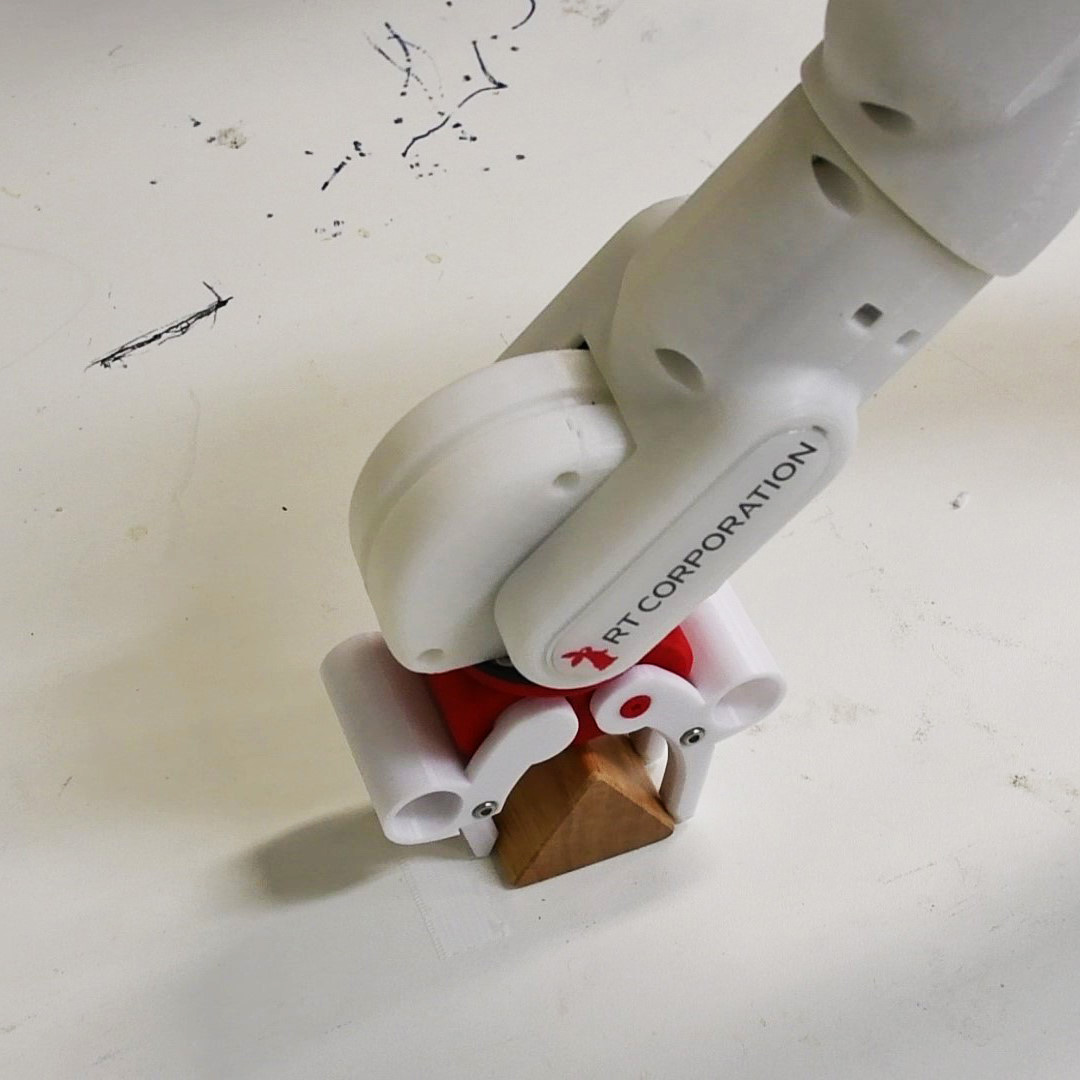}
        \subcaption{triangle block picking} 
    \end{minipage}
    \begin{minipage}[t]{0.49\linewidth}
        \centering
        \includegraphics[width=\linewidth]{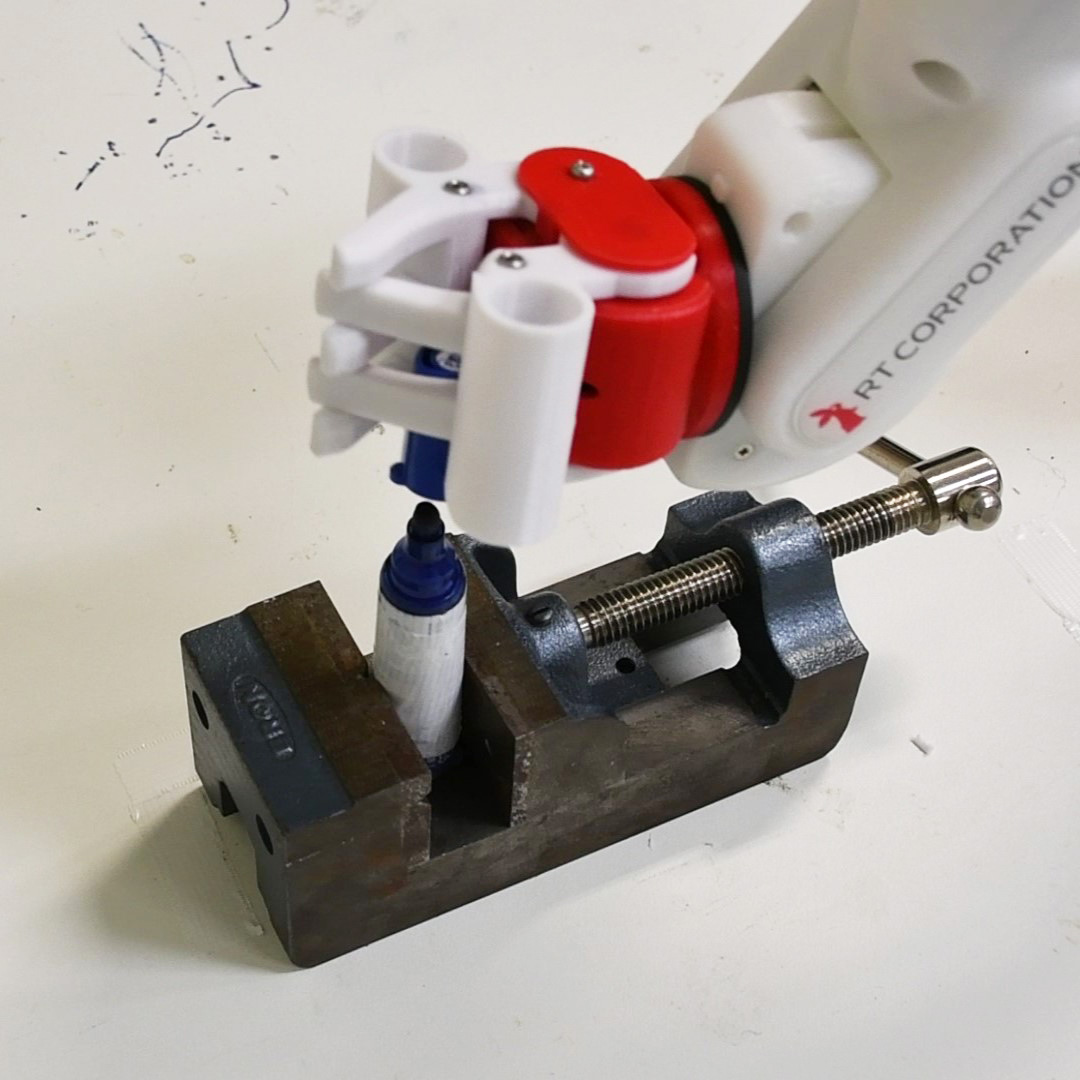}
        \subcaption{pen cap picking} 
    \end{minipage}
    \caption{Task for preliminary experiment}
    \label{fig:hand_test_tasks_overview}
\end{figure}

\begin{figure*}[tbp]
    
    \begin{minipage}[b]{0.19\linewidth}
        \centering
        \includegraphics[width=\linewidth]{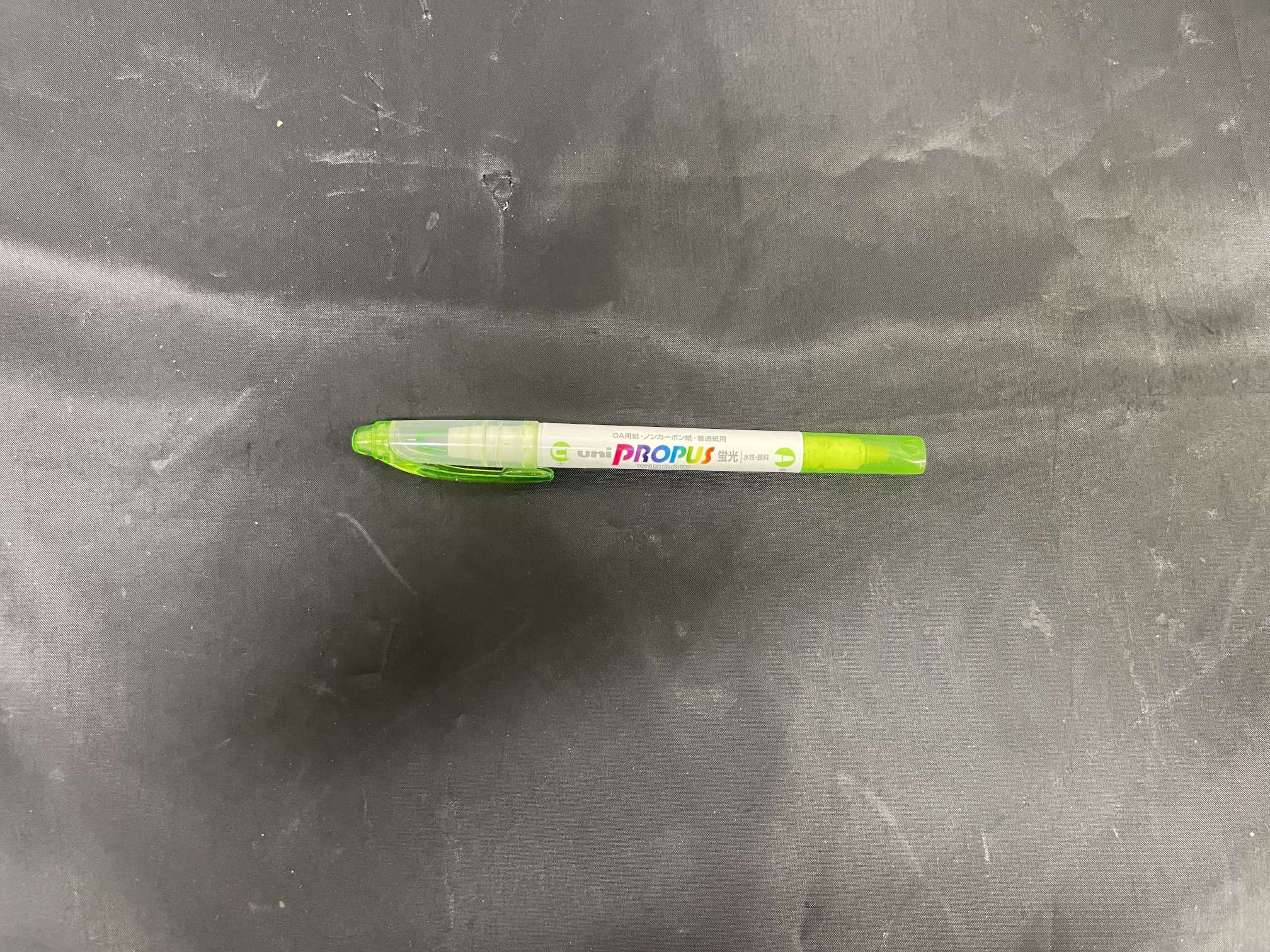}
        \subcaption{pen} 
    \end{minipage}
    \begin{minipage}[b]{0.19\linewidth}
        \centering
        \includegraphics[width=\linewidth]{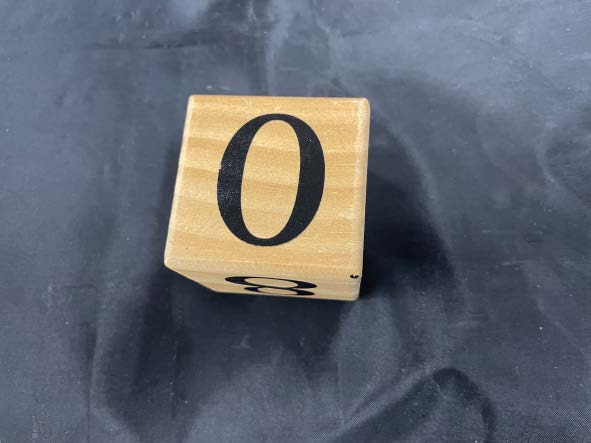}
        \subcaption{block} 
    \end{minipage}
    \begin{minipage}[b]{0.19\linewidth}
        \centering
        \includegraphics[width=\linewidth]{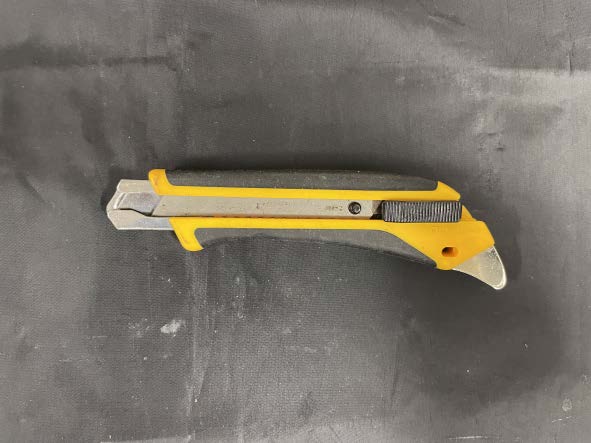}
        \subcaption{cutter} 
    \end{minipage}
    \begin{minipage}[b]{0.19\linewidth}
        \centering
        \includegraphics[width=\linewidth]{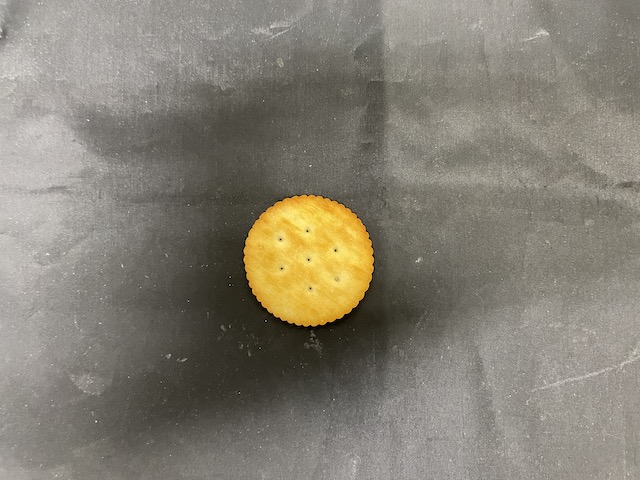}
        \subcaption{cracker}
    \end{minipage}
    \begin{minipage}[b]{0.19\linewidth}
        \centering
        \includegraphics[width=\linewidth]{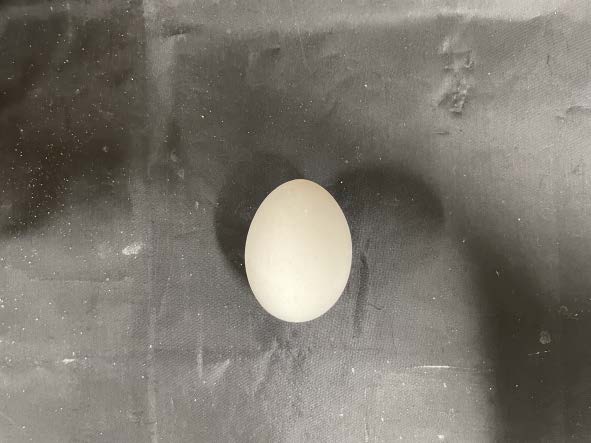}
        \subcaption{egg} 
    \end{minipage} \\
    
    \begin{minipage}[b]{0.19\linewidth}
        \centering
        \includegraphics[width=\linewidth]{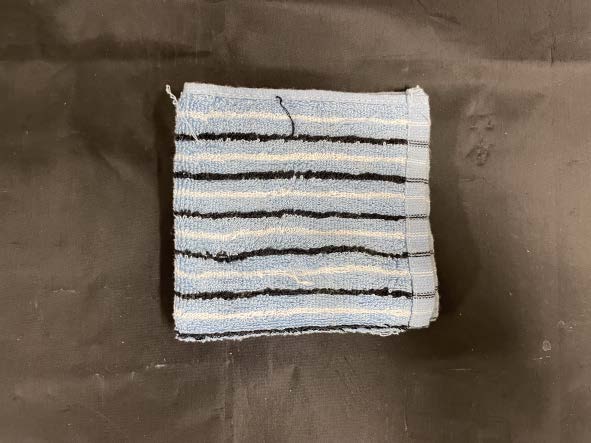}
        \subcaption{cloth}
    \end{minipage}
    \begin{minipage}[b]{0.19\linewidth}
        \centering
        \includegraphics[width=\linewidth]{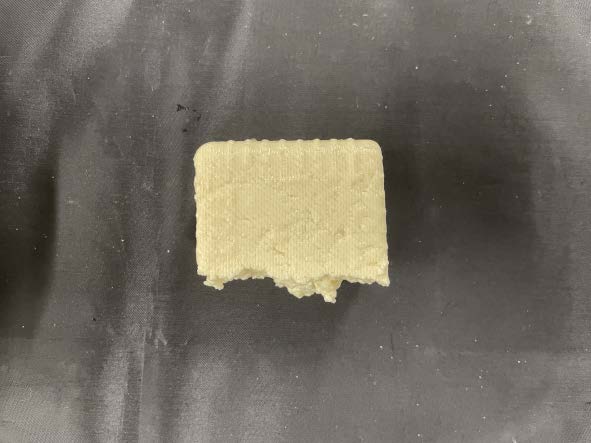}
        \subcaption{tofu}
    \end{minipage}
    \begin{minipage}[b]{0.19\linewidth}
        \centering
        \includegraphics[width=\linewidth]{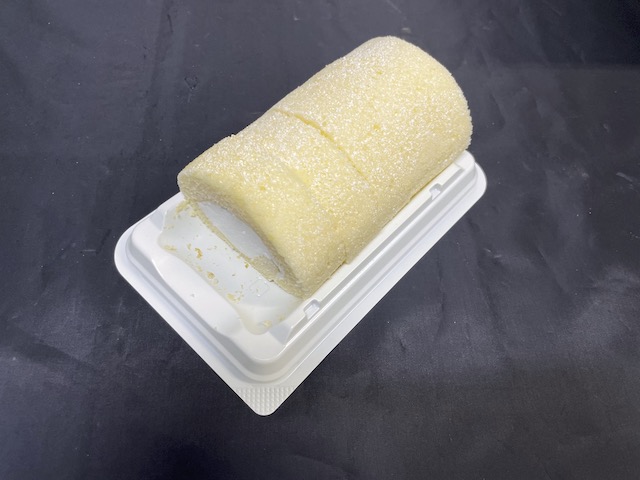}
        \subcaption{roll cake}
    \end{minipage}
    \begin{minipage}[b]{0.19\linewidth}
        \centering
        \includegraphics[width=\linewidth]{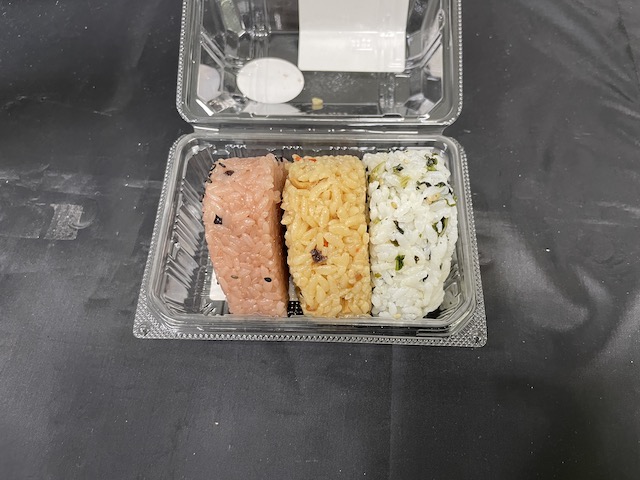}
        \subcaption{rice ball}
    \end{minipage}
    \begin{minipage}[b]{0.19\linewidth}
        \centering
        \includegraphics[width=\linewidth]{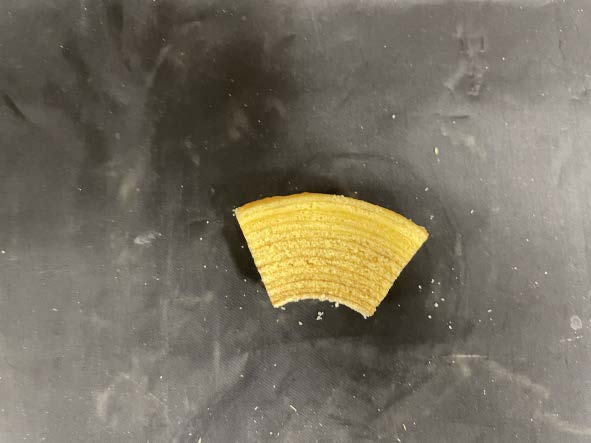}
        \subcaption{baumkuchen}
    \end{minipage} \\ 
    
    \begin{minipage}[b]{0.19\linewidth}
        \centering
        \includegraphics[width=\linewidth]{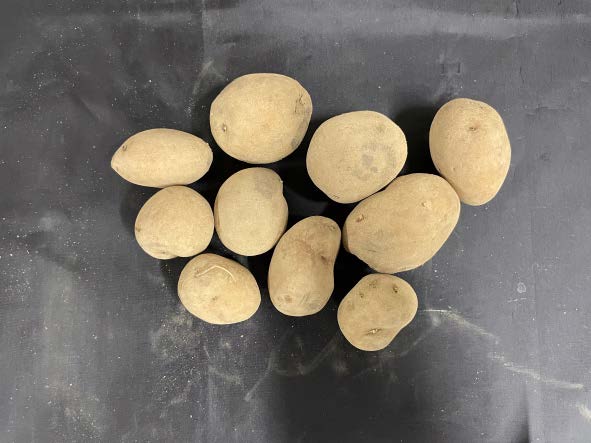}
        \subcaption{potato}
    \end{minipage}
    \begin{minipage}[b]{0.19\linewidth}
        \centering
        \includegraphics[width=\linewidth]{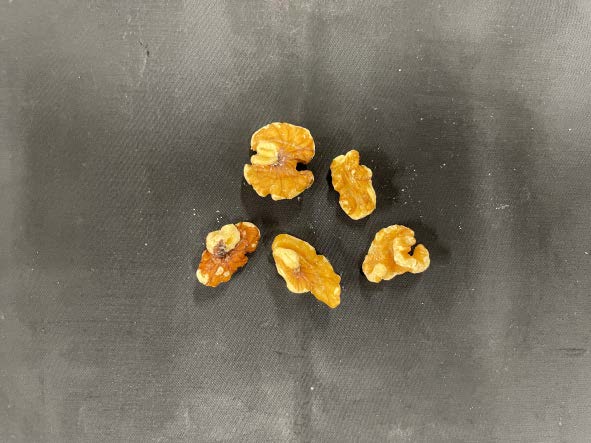}
        \subcaption{nuts}
    \end{minipage}
    \begin{minipage}[b]{0.19\linewidth}
        \centering
        \includegraphics[width=\linewidth]{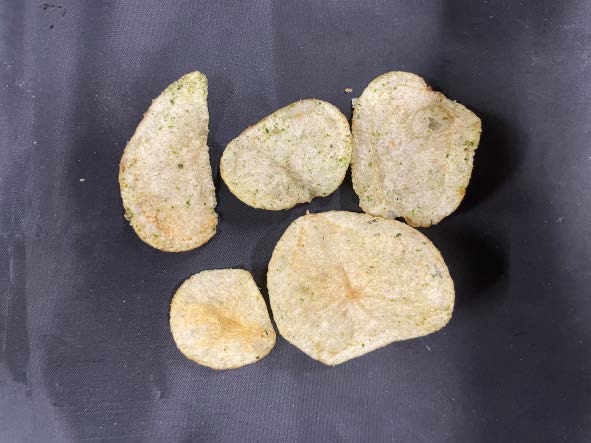}
        \subcaption{potato chips}
    \end{minipage}
    \begin{minipage}[b]{0.19\linewidth}
        \centering
        \includegraphics[width=\linewidth]{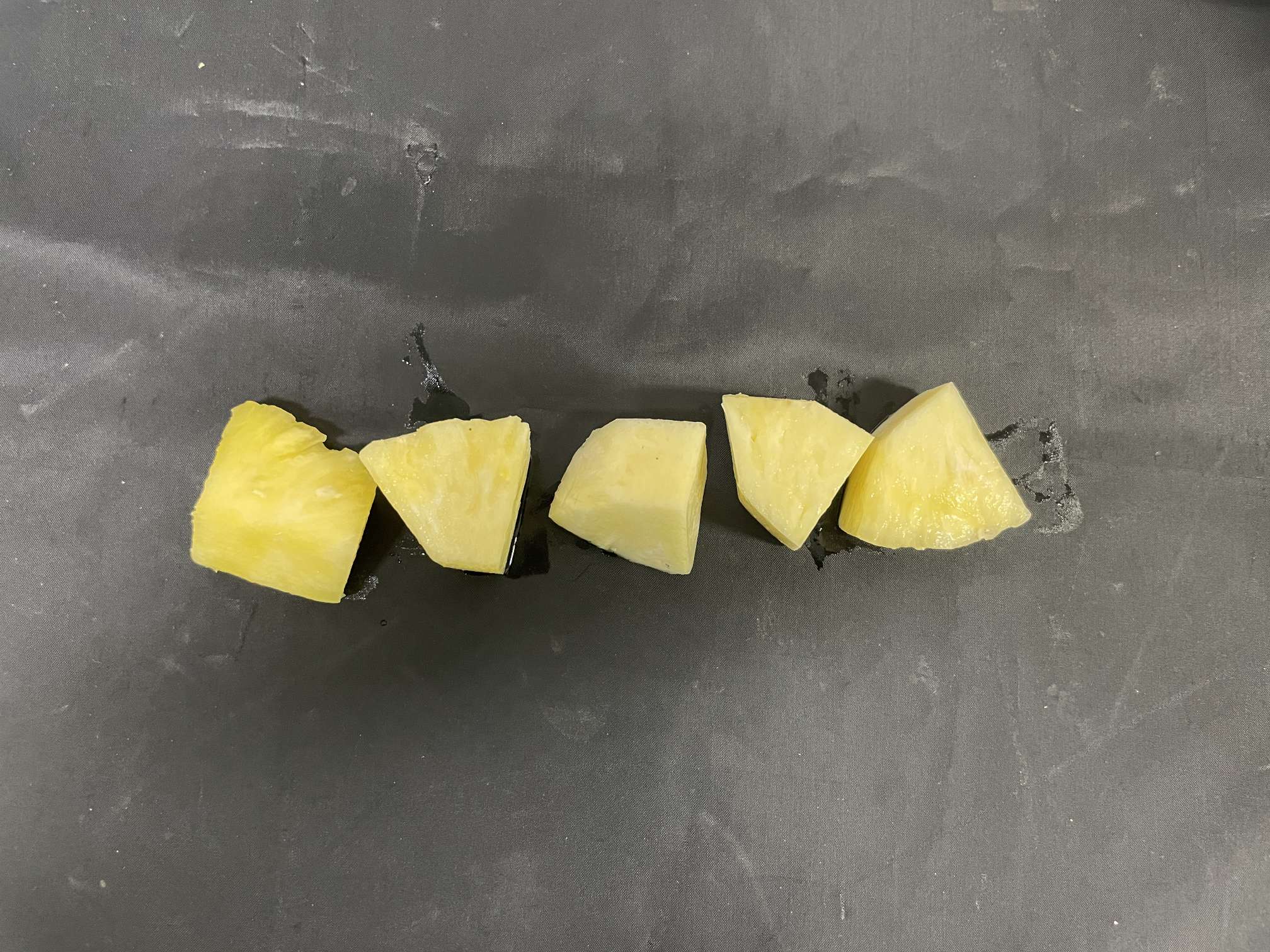}
        \subcaption{pineapple}
    \end{minipage}
    \begin{minipage}[b]{0.19\linewidth}
        \centering
        \includegraphics[width=\linewidth]{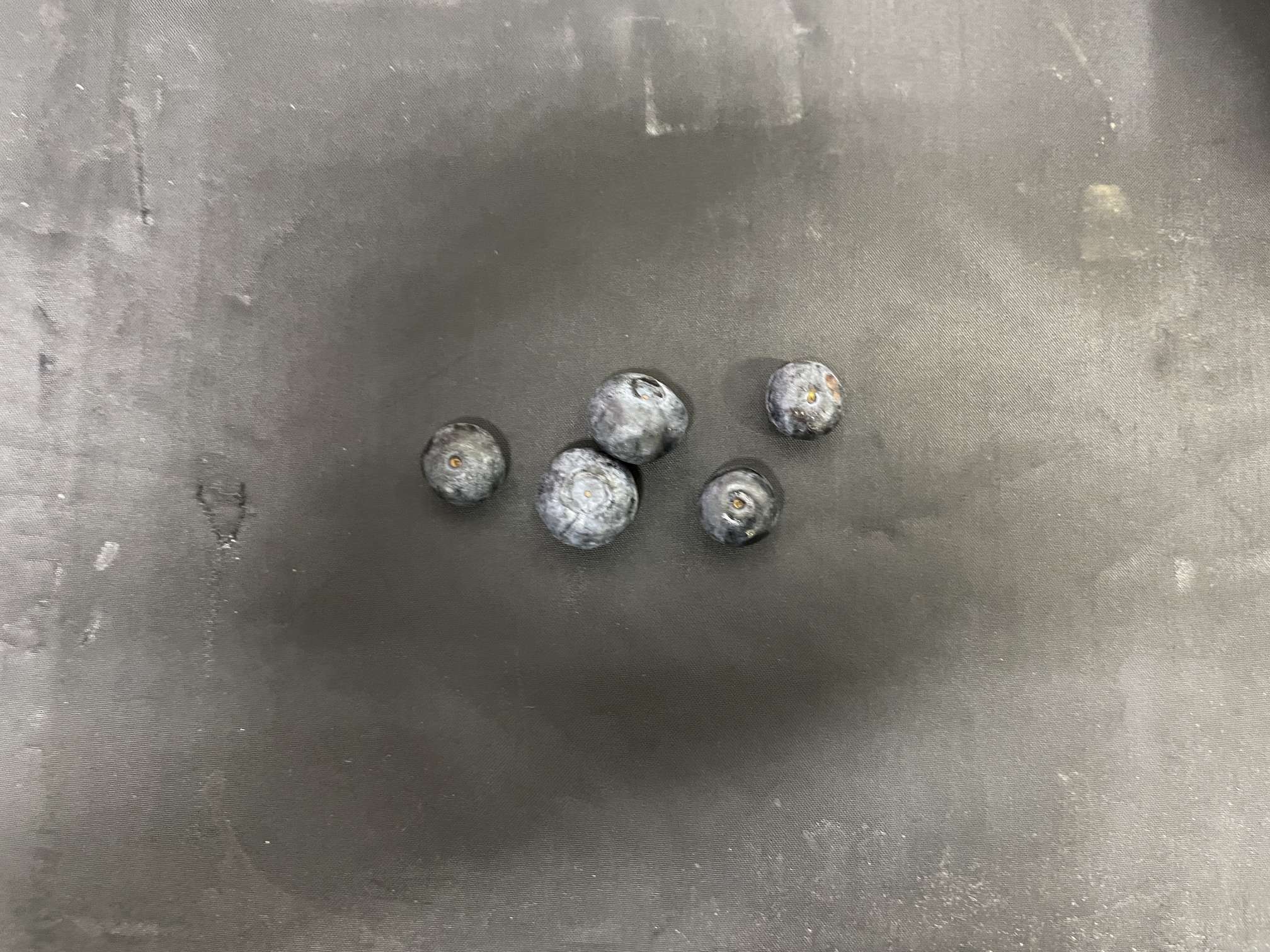}
        \subcaption{blueberry}
    \end{minipage} \\
    
    \begin{minipage}[b]{0.19\linewidth}
        \centering
        \includegraphics[width=\linewidth]{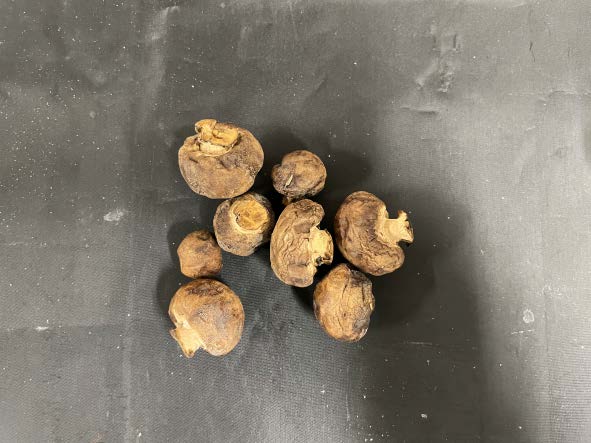}
        \subcaption{mushroom}
    \end{minipage}
    \begin{minipage}[b]{0.19\linewidth}
        \centering
        \includegraphics[width=\linewidth]{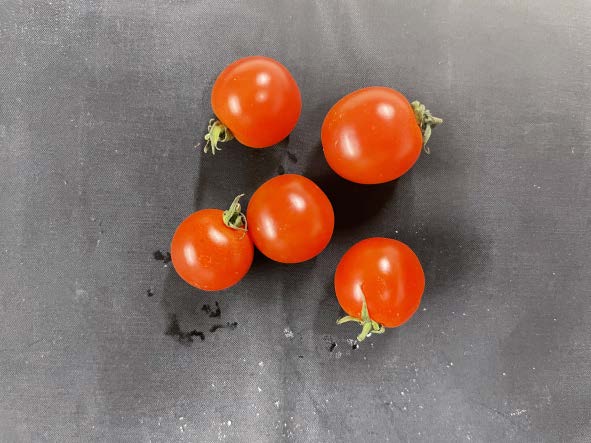}
        \subcaption{mini tomato}
    \end{minipage}
    \begin{minipage}[b]{0.19\linewidth}
        \centering
        \includegraphics[width=\linewidth]{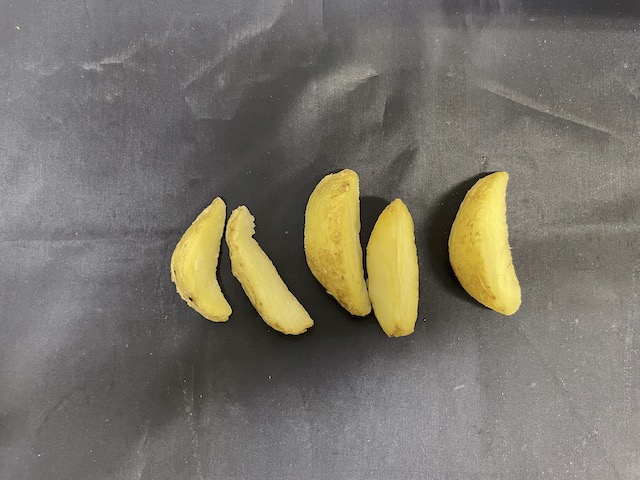}
        \subcaption{fried potato}
    \end{minipage}
    \begin{minipage}[b]{0.19\linewidth}
        \centering
        \includegraphics[width=\linewidth]{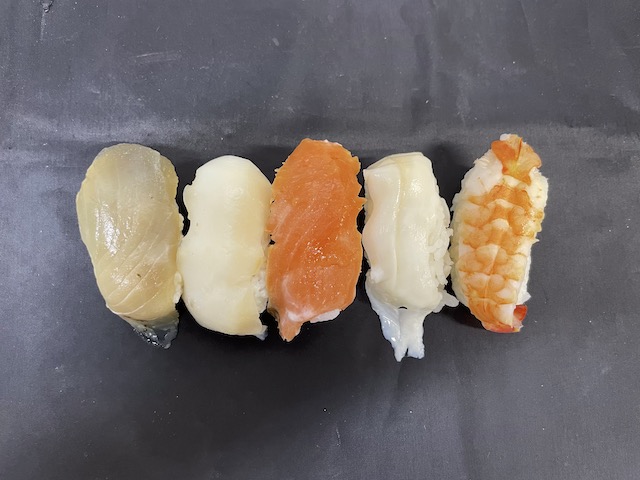}
        \subcaption{sushi}
    \end{minipage}
    \begin{minipage}[b]{0.19\linewidth}
        \centering
        \includegraphics[width=\linewidth]{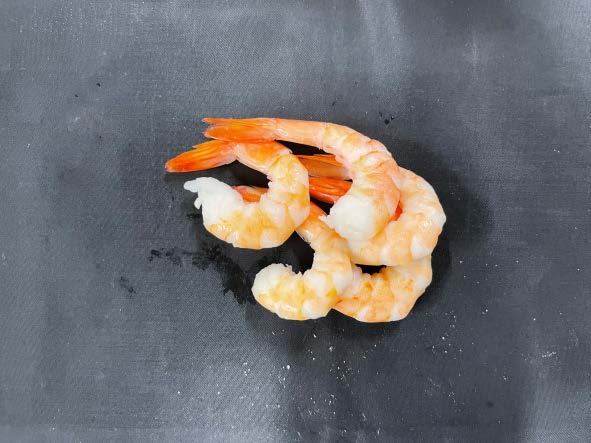}
        \subcaption{shrimp}
    \end{minipage}
    
    \caption{Objects used in the experiment}
    \label{fig:object}
\end{figure*}

\section{Experiment}

Two tasks were set up to evaluate the generalizability of soft and rigid grasping: a pick-and-place task as soft grasping and a letter-writing task as rigid grasping.

\subsection{Preliminary}
To confirm the effectiveness of the proposed hand, we created versions of the hand without the crossing structure and the claw, respectively.
To compare these hands, two typical grasping tasks are prepared: triangle block picking and pen cap picking.
The success rate was verified by teleoperation with 4-channel bilateral control, not imitation learning.
Experiments are conducted 5 times for each setting by one operator.
The hands for compare-experiments are shown in Fig.~\ref{fig:hands_for_preliminary_experiments}, the overviews of tasks are shown in Fig.~\ref{fig:hand_test_tasks_overview} and the results of the experiments are provided in Table~\ref{table:success_rate_hand}.

\subsubsection{Triangle block picking}
The first task was picking a triangle block, as shown in Figure ~\ref{fig:hand_test_tasks_overview} (a).
Success or failure was judged when the block completely left the table.
The results of the experiment showed that the hands without nails could not pick up.
It is because they cannot slide their fingers under the block.
On the other hand, the hands with claws were able to slide their fingertips under the block, and by enclosing the block with their hands, they were able to apply force appropriately and pick up it easily.
These results indicate that the fingertips need to be slid under the object to grasp an object with sloping sides, such as a triangular prism, which exerts downward force when a lateral force is applied, and that the nail-like shape is effective in such cases.

\subsubsection{Pen cap picking}
The second task is picking a pen cap, which involves grasping a thin, cylindrical object horizontally, as shown in Figure ~\ref{fig:hand_test_tasks_overview} (b).
Success or failure was judged when the cap was picked above the tip of the pen.
In the experiment, the hands with the cross-structure succeeded at all, whereas the hands without the cross-structure did not succeed at all.
To grasp a thin object with the one-degree-of-freedom rotary hand without the cross-structure, it is necessary to pinch the object with the tips of the fingers, which requires considerable accuracy of motion.
Even if the object can be grasped, the hand cannot be geometrically constrained because there are only two points of contact between the hand and the object.
On the other hand, a hand with a cross-structure can close without gaps, so that even a thin object can be grasped with the entire inside of the fingers.
Because of its adaptability, it is possible to grasp the object even if the positioning is not precise.
Moreover, it is possible to completely geometrically constrain the object by making three contact points.
This facilitates the task that an external force is applied to the grasped object such as reinserting the gripped cap into the pen or letter writing.

\begin{table}[tbp]
    \caption{Success rate of preliminary experiments}
    \label{table:success_rate_hand}
    \centering
    \begin{tabular}{c|cccc}
     \hline
     task & proposed & w/o nail & w/o cross  & \makecell{w/o nail \\ w/o cross} \\
     \hline
     triangle block picking & $\mathbf{100}$ & 0 & $\mathbf{100}$  & 0 \\
     pen cap picking & $\mathbf{100}$ & $\mathbf{100}$ & 0 & 0 \\
     \hline
    \end{tabular}
\end{table}

\subsection{Pick-and-Place Task}

\subsubsection{Task Design}

\begin{figure*}[t]
    \centering
    \begin{minipage}[t]{0.24\linewidth}
        \centering
        \includegraphics[width=\linewidth]{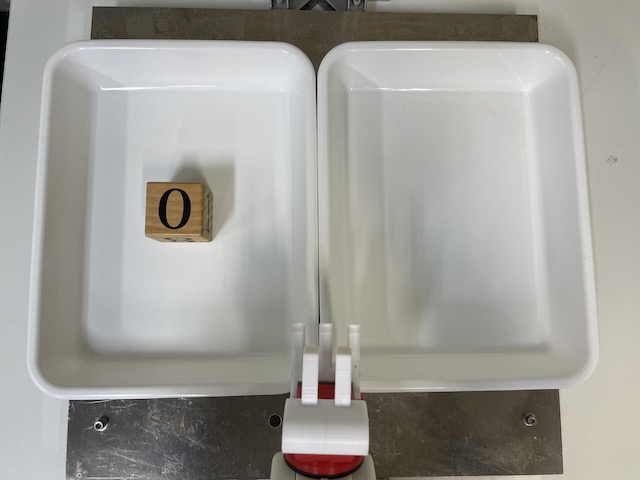}
        \subcaption{ENV of pick-and-place task}
        \label{fig:environment}
    \end{minipage}
    \begin{minipage}[t]{0.24\linewidth}
        \centering
        \includegraphics[width=\linewidth]{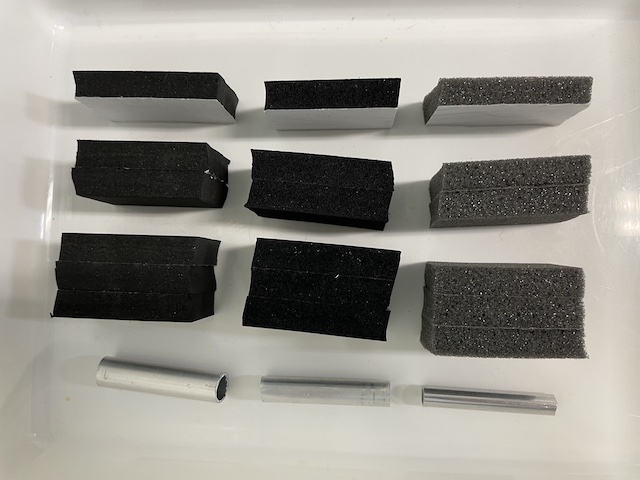}
        \subcaption{Objects for training}
        \label{fig:trained-objects}
    \end{minipage}
    \begin{minipage}[t]{0.24\linewidth}
        \centering
        \includegraphics[width=\linewidth]{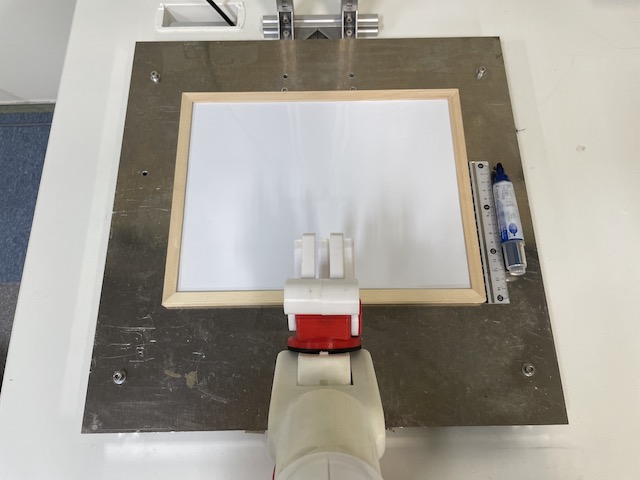}
        \subcaption{ENV of letter-writing task}
        \label{fig:environment-letter-writing}
    \end{minipage}
    \begin{minipage}[t]{0.18\linewidth}
        \centering
        \includegraphics[width=\linewidth]{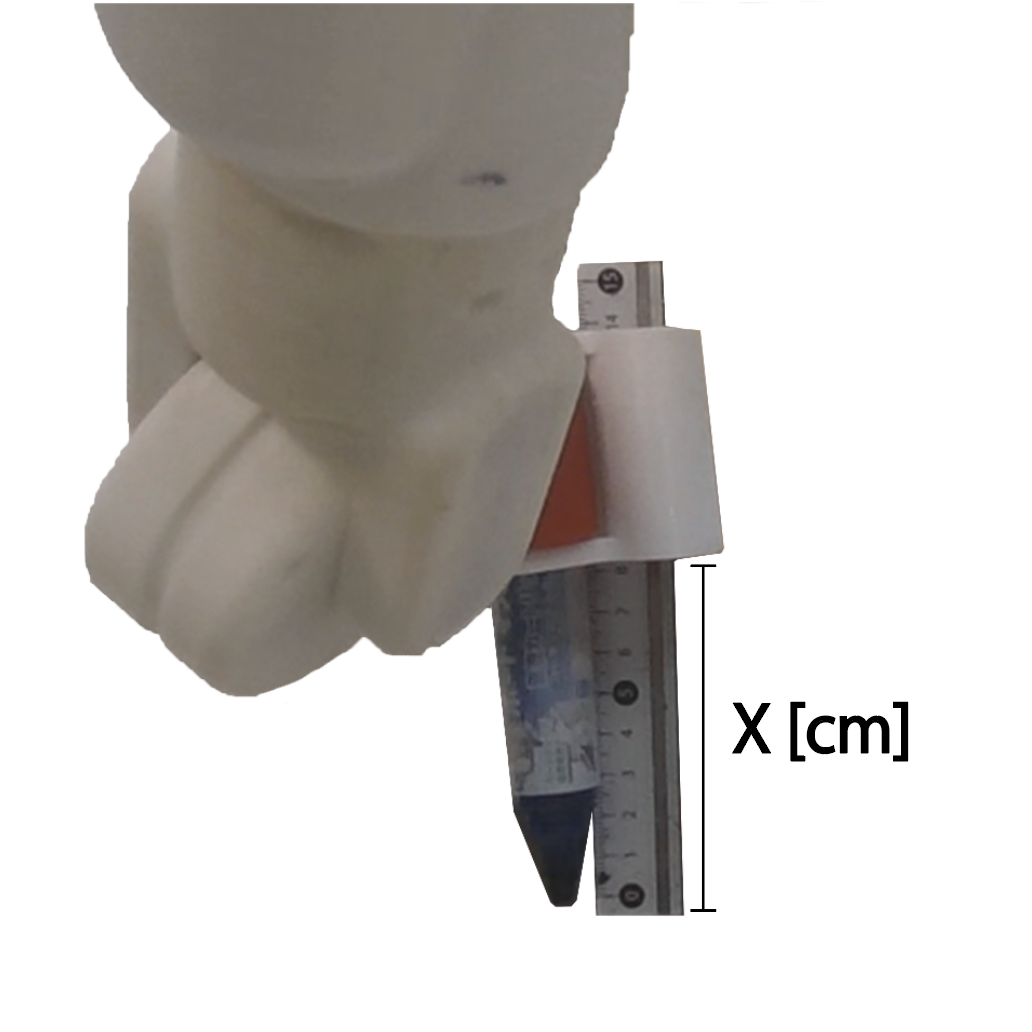}
        \subcaption{Positions of the pen}
        \label{fig:pen-position}
    \end{minipage}
    \caption{Experimental setups}
    \label{fig:experimental_setups}
\end{figure*}

An overview of the environment considered in this experiment is presented in Fig.~\ref{fig:experimental_setups}~(a).
The robot picked the object at the center of the left tray and placed it at the right tray's center.
Demonstration data were collected using 12 objects, and we collected 2 training datasets and 1 validation dataset per object.
Images of the objects used to collect the demonstration data are presented in Fig.~\ref{fig:experimental_setups}~(b).
Trained objects include 9 sponges and 3 aluminum pipes.
The width of each sponge was 10, 20, and 30 mm, and the constituent materials were ethylene propylene diene monomer (EPDM), natural rubber (NR), and urethane, respectively, which were low-cost and high-availability materials.
Furthermore, each aluminum pipe had a different diameter: 8, 10, and 12 mm, respectively.
To collect demonstration data, we applied scaling bilateral control, scales the force of the follower's hand to 1/10 that of the leader.
By employing scaling bilateral control, higher precision motion than that of a human can be obtained.
For imitation learning, a 6-layer LSTM was employed.
Adam~\cite{kingma2014adam} was utilized for optimization, and 10000 epochs were trained with a batch size of 16.
In total, 20 types of objects were employed in the experiment to test autonomous control.
Experiments are conducted 5 times for each object.
Images of the objects utilized in the experiment are presented in Fig.~\ref{fig:object}.
The tasks were tested with force control as the proposed method (F2FL) and without force control but only position control as the baseline method, i.e., $\bm{K}_f=0$ in Eq. (1) and using the same LSTM model as that of the proposed method (F2FL-w/o-Force).

\begin{table}[tbp]

    \caption{Success rate of pick-and-place task}
    \label{table:success_rate}
    \centering
    \begin{tabular}{c|c|cc}
    \hline
    \multicolumn{2}{c|}{Test object} & \multicolumn{2}{c}{Success rate} \\
    \hline
    \multicolumn{1}{c|}{name} & weight [g] & F2FL (proposed) & F2FL-w/o-Force \\
    \hline
    \hline
    pen          &              7.8 &             60 &              0 \\
    block        &             50.9 & $\mathbf{100}$ &             80 \\
    cutter       &             77.8 & $\mathbf{100}$ &             40 \\
    cracker      &   3.4 $\sim$ 3.5 & $\mathbf{100}$ &              0 \\
    egg          & 49.1 $\sim$ 54.4 & $\mathbf{100}$ &              0 \\
    cloth        &             28.0 &             80 &             40 \\
    baumkuchen   & 21.7 $\sim$ 21.9 & $\mathbf{100}$ &             60 \\
    rice ball    & 73.9 $\sim$ 75.1 & $\mathbf{100}$ &             60 \\
    roll cake    & 35.9 $\sim$ 38.8 & $\mathbf{100}$ &             20 \\
    tofu         & 60.0 $\sim$ 70.0 & $\mathbf{100}$ & $\mathbf{100}$ \\
    potato       & 93.7 $\sim$ 131.1 & $\mathbf{100}$ &             40 \\
    nuts         &    0.9 $\sim$ 2.8 & $\mathbf{100}$ &             20 \\
    potato chips &    0.8 $\sim$ 1.8 & $\mathbf{100}$ &             20 \\
    pineapple    &  14.6 $\sim$ 23.4 & $\mathbf{100}$ &             60 \\
    blueberry    &    1.1 $\sim$ 2.4 &             60 & $\mathbf{100}$ \\
    mushroom     &  5.0 $\sim$ 22.7 & $\mathbf{100}$ &             60 \\
    mini tomato  & 11.1 $\sim$ 22.3 & $\mathbf{100}$ &             60 \\
    fried potato &  2.7 $\sim$ 22.9 & $\mathbf{100}$ &             80 \\
    sushi        & 23.1 $\sim$ 31.2 & $\mathbf{100}$ &             80 \\
    shrimp       &   5.1 $\sim$ 6.8 & $\mathbf{100}$ &             60 \\
    \hline
    \multicolumn{2}{c|}{Total} & $\mathbf{95}$ & 49 \\
    \hline
    \end{tabular}

\end{table}

\begin{figure*}[t]
    \centering
    \begin{minipage}[t]{0.49\linewidth}
        \centering
        \includegraphics[width=\linewidth]{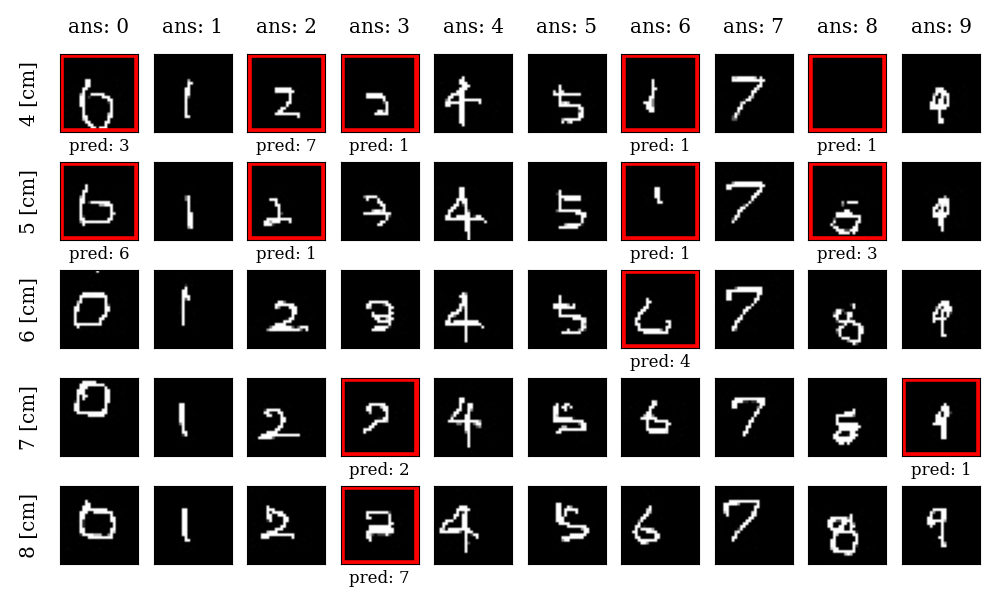}
        \subcaption{F2FL (proposed method)}
    \end{minipage}
    \begin{minipage}[t]{0.49\linewidth}
        \centering
        \includegraphics[width=\linewidth]{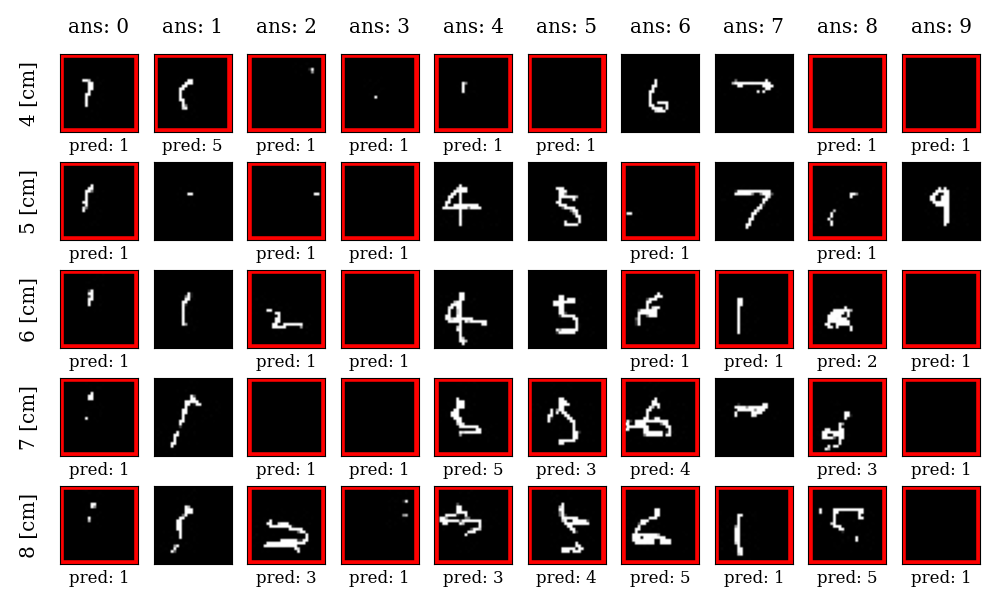}
        \subcaption{F2FL-w/o-force}
    \end{minipage}
    \caption{Results of writing task}
    \label{fig:results_writing}
\end{figure*}

\subsubsection{Results}

The success rate of the pick-and-place task of F2FL and F2FL-w/o-Force are presented in Table \ref{table:success_rate}.

Regarding F2FL, it was demonstrated that grasping motion is possible with a high success rate, even for objects whose shape and hardness were not known in advance.
We also verified that the robot could grasp a low-stiffness object such as tofu with appropriate force to accomplish the task.
One failure instance occurred when the pen's center of gravity was biased, causing the pen to be dropped because the force of the learned grasping motion was weak, the shape of the pen was long and slender, and its surface was slippery.
In addition, when gripping a cloth, if the cloth was flat, there was no part to be caught, and it could not be gripped. 
Furthermore, the smallest object, namely a blueberry, sometimes fell between the fingers of the hand.
The finger space of the hand was 10.4 mm, and the blueberry was the only object that included a smaller one than this space.
This result suggests limitations of the cross-structure hand.

However, for the F2FL-w/o-Force, several mistakes were observed for various objects.
In particular, for the pen, which had the narrowest width, several failures were registered.
In addition, the cracker and egg, which presented large widths, could be easily damaged.
The experiments with these objects were totally unsuccessful owing to torque limitations or because the objects crushed.
Easily collapsible objects, such as tofu and sushi, whose widths were close to those of the learning objects, exhibited a high success rate.
Also, blueberry exhibited a higher success rate than the proposed method. 
It is because in some cases blueberries are wet and stuck to the fingers without force.

These results indicate that when only position control was utilized, the robot succeeded when the object's width was close to that of the learned object but failed otherwise.
In contrast, when force control was employed, the error between the position command value and object width was compensated by force control, and a high success rate was achieved for several objects.

\subsection{Letter-Writing Task}

\subsubsection{Task Design}
The writing test was conducted because such a task requires a tool.
This task involved writing a letter containing numbers from zero to nine on a whiteboard using a whiteboard marker.
The overall environment considered in this experiment is depicted in Fig.~\ref{fig:experimental_setups}~(c).

In total, six demonstration datasets were collected for each number twice in each of the three settings.
These settings involve different pen arrangements.
Each setting shifted the pen placement by 1 cm along the ruler.
Each set was named according to the length from the hand to the pen tip: 4, 5, 6, 7, and 8 cm, respectively.
The length was measured, as shown in Fig.~\ref{fig:experimental_setups}~(d).
For imitation learning, 8-layer LSTM and data collected in the 4, 6, and 8 cm settings were adopted.

The evaluation was conducted in five environments: three trained settings (4, 6, and 8 cm) and two untrained settings (5, 7 cm), once per setting.
Each trial was checked via image classification.
ResNet18~\cite{he2016deep} was employed for classification; it was trained using MNIST dataset~\cite{lecun1998gradient} and the validation success rate was 99.6\%.
In addition, to verify the need for force control, trials were also conducted using the F2FL-w/o-force model.



\subsubsection{Results}
The results obtained from the proposed method are presented in Fig.~\ref{fig:results_writing}~(a), whereas the results from adopting only position control (F2FL-w/o-force) are presented in Fig.~\ref{fig:results_writing}~(b).
In Fig.~\ref{fig:results_writing}, ``ans'' denotes the correct answer to the classification, whereas ``pred'' represents the classification result in failure cases.
The area enclosed in red represents the failure case.
The vertical axis represents the position of the pen.
The total success rates of the proposed method and F2FL-w/o-force were 74\% and 26\%, respectively.

For the failure cases of the proposed method, many failures were observed in letters featuring many curved lines, such as 3, 6, and 8.
It was difficult to apply strong pressure to letters with many curves, especially when the pen was held short, and it was inferred that the pen tended to float.

Regarding position control only (F2FL-w/o-force), there were many cases where the pen floated; in these cases, excessive force was applied at the beginning and the pen was not applied properly, or the pen was dropped and the characters could not be written.
The successful cases were often intermediate cases that involved holding the pen at 5 cm, 6 cm, etc. and were considered to be outputting the average behavior of the training data.

These results indicate that it is possible to learn tool-based tasks using imitation learning based on the developed hand and bilateral control; in addition, the effectiveness of employing force control was verified.

\section{CONCLUSIONS}

In this study, we verified that the proposed cross-structure hand can measure the force of a human's motion in complex grasping tasks.
Moreover, we demonstrated that the robot can perform soft and rigid grasping.
As a soft grasping task, the pick-and-place task, including non-rigid and irregular objects, was tested.
The method was applied to several objects, including non-rigid and irregular objects.
Furthermore, as a rigid grasping task, letter writing was tested.
Accordingly, the method succeeded in writing letters in several situations, including unlearned ones.
More specifically, we demonstrated that leveraging on high-quality teaching data collected using bilateral control and a hand capable of force control that can grasp both thick and thin objects, complicated motion planning can be achieved even with a simple LSTM.
Therefore, the proposed method is highly practical because it is easy to learn.

However, to achieve further complex tasks, a more sophisticated inference model, 
such as that in~\cite{zeng2021transporter}, will be essential.
Therefore, we will address these aspects in the future.

\section*{ACKNOWLEDGMENT}
This study is based on results obtained from a project, JPNP20004, subsidized by the New Energy and Industrial Technology Development Organization (NEDO) and also supported by the Japan Society for the Promotion of Science through a Grant-in-Aid for Scientific Research (B) under Grant 21H01347.

\bibliographystyle{ieeetr}
\bibliography{reference.bib}

\end{document}